\documentclass[10pt,twocolumn,letterpaper]{article}

\usepackage[pagenumbers]{iccv} % To force page numbers, e.g. for an arXiv version

\usepackage[most]{tcolorbox}

\newtcolorbox{disclaimerbox}{
    colback=gray!10,     % background color
    colframe=gray!40,    % frame color
    boxrule=0.5pt,       % border thickness
    arc=4pt,             % rounded corners
    auto outer arc,
    boxsep=5pt,
    left=6pt,
    right=6pt,
    top=4pt,
    bottom=4pt,
    enhanced jigsaw
}

\definecolor{iccvblue}{rgb}{0.21,0.49,0.74}
\usepackage[pagebackref,breaklinks,colorlinks,allcolors=iccvblue]{hyperref}

\def\paperID{*****} % *** Enter the Paper ID here
\def\confName{ICCV}
\def\confYear{2025}

\title{Efficient Real-World Deblurring using Single Images: \\ AIM 2025 Challenge Report}

\author{
Daniel Feijoo$^{1}$\thanks{Corresponding authors: {\tt\small \{marcos.conde, danfei\}@cidaut.es}.} \quad
Paula Garrido\mbox{-}Mellado$^{1}$ \quad
Marcos V.\ Conde$^{1,3}$ \quad
Jaesung Rim$^{2}$ \\
Álvaro García$^{1}$ \quad
Sunghyun Cho$^{2}$ \quad
Radu Timofte$^{3}$ \\ \\
$^{1}$\,Cidaut AI, Spain \quad
$^{2}$\,POSTECH, Korea \quad
$^{3}$\,University of W\"urzburg, Germany
}

\begin{document}
\maketitle

\begin{abstract}
% This abstract is partially based in the work 
% NTIRE 2025 Challenge on Efficient Burst HDR and Restoration:
%Datasets, Methods, and Results
This paper reviews the AIM 2025 Efficient Real-World Deblurring using Single Images Challenge, which aims to advance in efficient real-blur restoration. The challenge is based on a new test set based on the well known RSBlur dataset. Pairs of blur and degraded images in this dataset are captured using a double-camera system. Participant were tasked with developing solutions to effectively deblur these type of images while fulfilling strict efficiency constraints: fewer than 5 million model parameters and a computational budget under 200 GMACs. A total of 71 participants registered, with 4 teams finally submitting valid solutions. The top-performing approach achieved a PSNR of 31.1298 dB, showcasing the potential of efficient methods in this domain. This paper provides a comprehensive overview of the challenge, compares the proposed solutions, and serves as a valuable reference for researchers in efficient real-world image deblurring. 
\end{abstract}

\section{Introduction}
\label{sec:intro}

Motion blur often occurs in low-light environments when longer exposures are required, degrading perceptual quality and hampering downstream vision tasks in imaging systems. 
The blur arises more frequently on smartphones because their small sensors collect less light and therefore necessitate longer shutter times. Consequently, there is growing interest in deblurring methods that can run efficiently on mobile devices.

Recently, learning-based deblurring methods have been widely studied. By training on large datasets, these methods learn a direct mapping from a blurred input to its latent sharp image and have achieved superior improvement in restoration quality. 
Advances in sophisticated architectures~\cite{Cho2021,Zamir2021Restormer,sfhformer,chen2022simple,tsai2022banet,tsai2022stripformer,kong2023efficient, Feijoo_2025_CVPR} have further enhanced their ability to handle motion blur, pushing the state-of-the-art performance on deblurring benchmarks. 
Despite this progress, efficient deblurring on mobile remains challenging. 
Deblurring typically requires a large effective receptive field, which in turn drives up FLOPs, memory footprint, and latency.
As a result, computational cost and inference time remain substantial, limiting mobile applications.

%Requirement of processing high-resolution inputs and mataining a large effective receptive field for effective deblurring drive up FLOPs, memory footprint, and latency.

%Processing high-resolution inputs and the need for a large receptive field for deblurring drive up FLOPs, memory footprint, and latency. 

%디블러링 은 종종 large effective field를 필요로 하며 

%High-resolution inputs and the need for a large effective receptive field drive up FLOPs, memory footprint, and latency, making real-time inference difficult within the energy and thermal budgets of mobile processors.

%Recently, learning-based deblurring methods~\cite{} are widely studied. These methods learn the deblurring process from training datasets and have significantly improved performance.
%Advances in sophisticated architectures~\cite{} have further enhanced their ability to handle motion blur. 
%Despite these advances, computational cost and inference time remain substantial, limiting deployment on on-device and edge device applications.

%Learning-based deblurring methods requires large-scale datasets for training. 

For training learning-based methods, large-scale datasets are required. 
Most approaches rely on synthetic blur datasets~\cite{Nah2017,nah2019ntire,su2017deep} generated by averaging sharp images.
However, networks trained on such synthetic datasets often fail on real-world blurred images~\cite{rim2022realistic} because synthetic blur cannot reflect key factors of real images, including continuous motion, saturated pixels, sensor noise, and ISP effects.
To address this, several real-world blur datasets~\cite{rim2022realistic,realblur,zhong2020efficient,zhong2023real,Li2023} have been proposed,  collected with a dual-camera system. These works demonstrate that real-world datasets are essential for both training and evaluation in practical real-world deblurring.

Inspired by these, we propose a novel challenge in AIM 2025 to restore single blurred images. Compared to previous challenges~\cite{nah2019ntire, nah2019ntire_sr, nah2020ntire, nah2021ntire}, we introduced a more practical challenge by incorporating the real-world motion-blurred dataset. In addition, we establish some computational constraints on the proposed solutions. The computational constraints have been addressed by other image restoration challenges ~\cite{lee2025ntire, ren2024ninth, ren2025tenth, Zamfir_2023_CVPR, Conde_2023_CVPR}, but it has not been tackled by other deblurring challenges. The motivation for these computational constraints is to advance in blur restoration models that can be run in mobile or edge devices.

% Inspired by previous works~\cite{nah2019ntire, nah2019ntire_sr, nah2020ntire, nah2021ntire, abuolaim2021ntire}, we proposed a novel challenge in AIM 2025 to restore single blurred images. Compared to previous challenges, we introduced a more realistic challenge by incorporating real-world motion-blurred images. In addition, we establish some computational constraints on the proposed solutions. The computational constraints have been addressed by other different image restoration challenges ~\cite{lee2025ntire, ren2024ninth, ren2025tenth, khan2022ntire}, but it has not been tackled by other blur restoration challenges. The motivation for these computational constraints is to advance in blur restoration models that can be run in edge devices.

% This Challenge is one of the AIM 2025 Workshop associated challenges on: \modify{Add the different AIM Challenges}.

\section{AIM 2025 Efficient Real-World Deblurring using Single Images}
\label{sec:challenge}

The goal of this competition is to design an efficient algorithm for single-image deblurring. A simple baseline code and model are provided for participants as a starting kit (\url{https://github.com/cidautai/Efficient-Real-World-Deblurring}).

% \subsection{Overview}
% \label{subsec:overview}

\begin{figure}[t]
    \centering
    \includegraphics[width=0.9\columnwidth]{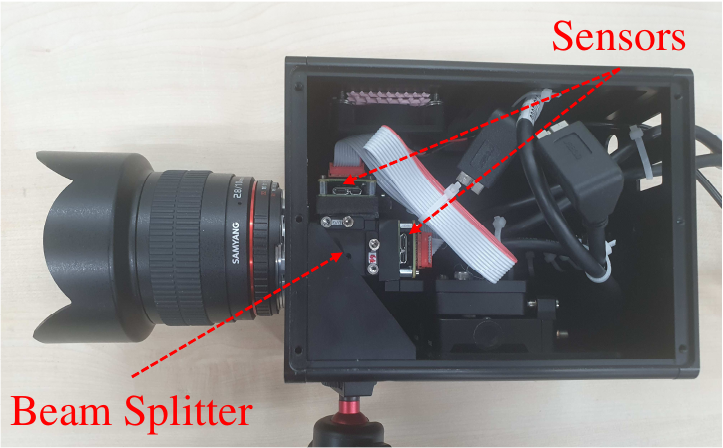}
    \caption{A dual-camera system for collecting the RSBlur dataset.}
    \label{fig:rsblur_system}
\end{figure}

\subsection{Datasets}
\label{subsec:datasets}
%For this challenge, an extension of the RSBlur~\cite{rim2022realistic} dataset was proposed to encourage the design of new algorithms for efficient and real-world blur restoration. 

For this challenge, the RSBlur~\cite{rim2022realistic} dataset is utilized to encourage the design of new algorithms for efficient and real-world blur restoration.

The RSBlur dataset was collected using a dual-camera system, as shown in \Cref{fig:rsblur_system}. 
The system consists of a single lens, a beam-splitter, and two imaging sensors. 
Light entering the lens is divided by the beam splitter and directed to both sensors. One sensor captures short-exposure images as ground-truth sharp images, while the other simultaneously captures a real-world blurred image with a long exposure time.
This capturing process yields a pair of aligned blur-sharp images.
The RSBlur dataset provides real-world blurred images and corresponding ground-truth images collected by using the dual-camera system. The dataset consists of 8,887, 1,120, and 3,360 paired images for training, validation, and test sets, respectively. 
We use the original RSBlur training and test sets for the development phase.

\begin{figure}[htbp]
    \centering
    \begin{subfigure}[b]{0.45\columnwidth}
        \includegraphics[width=\textwidth]{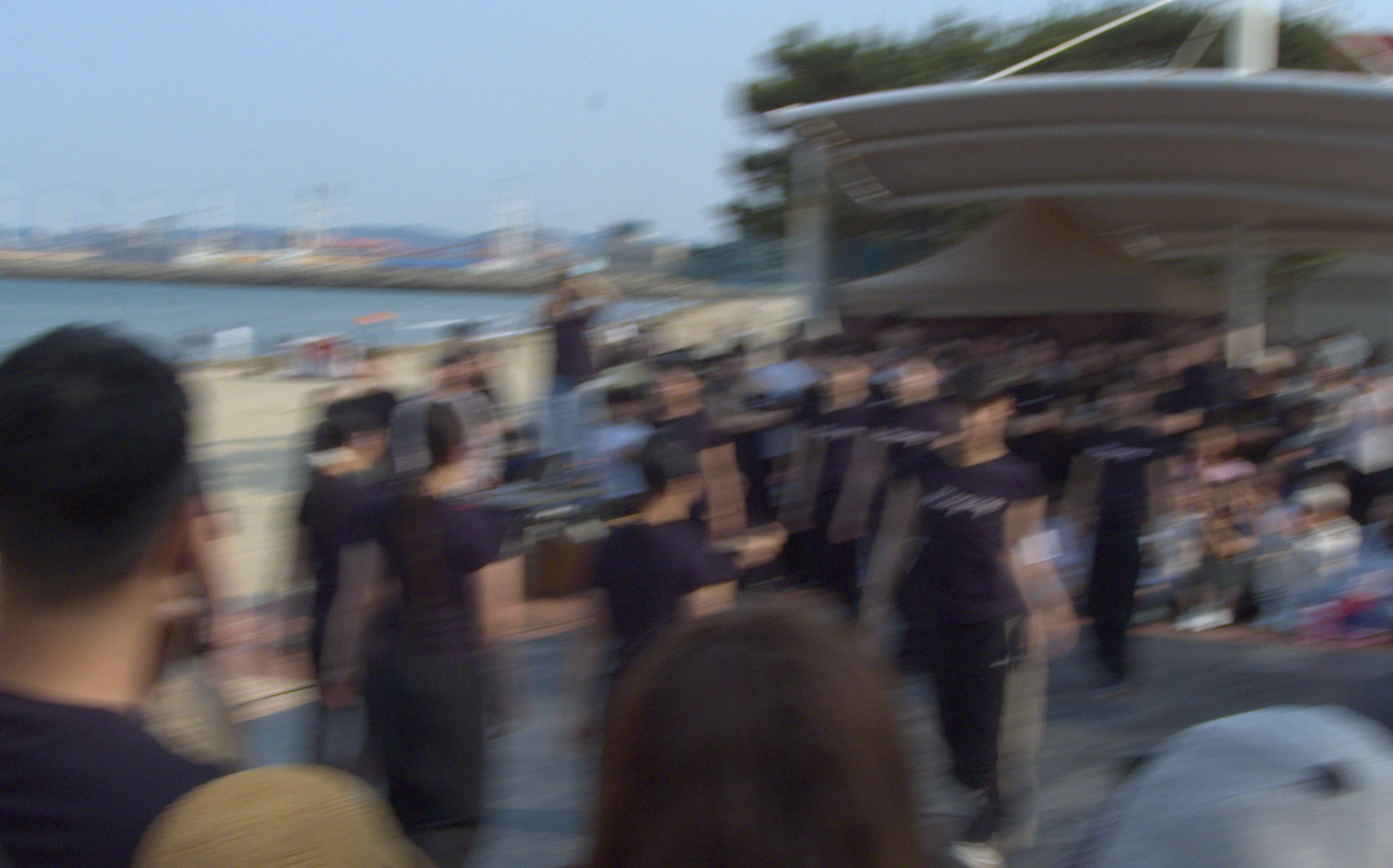}
        % \caption{Caption 1}
    \end{subfigure}
    \begin{subfigure}[b]{0.45\columnwidth}
        \includegraphics[width=\textwidth]{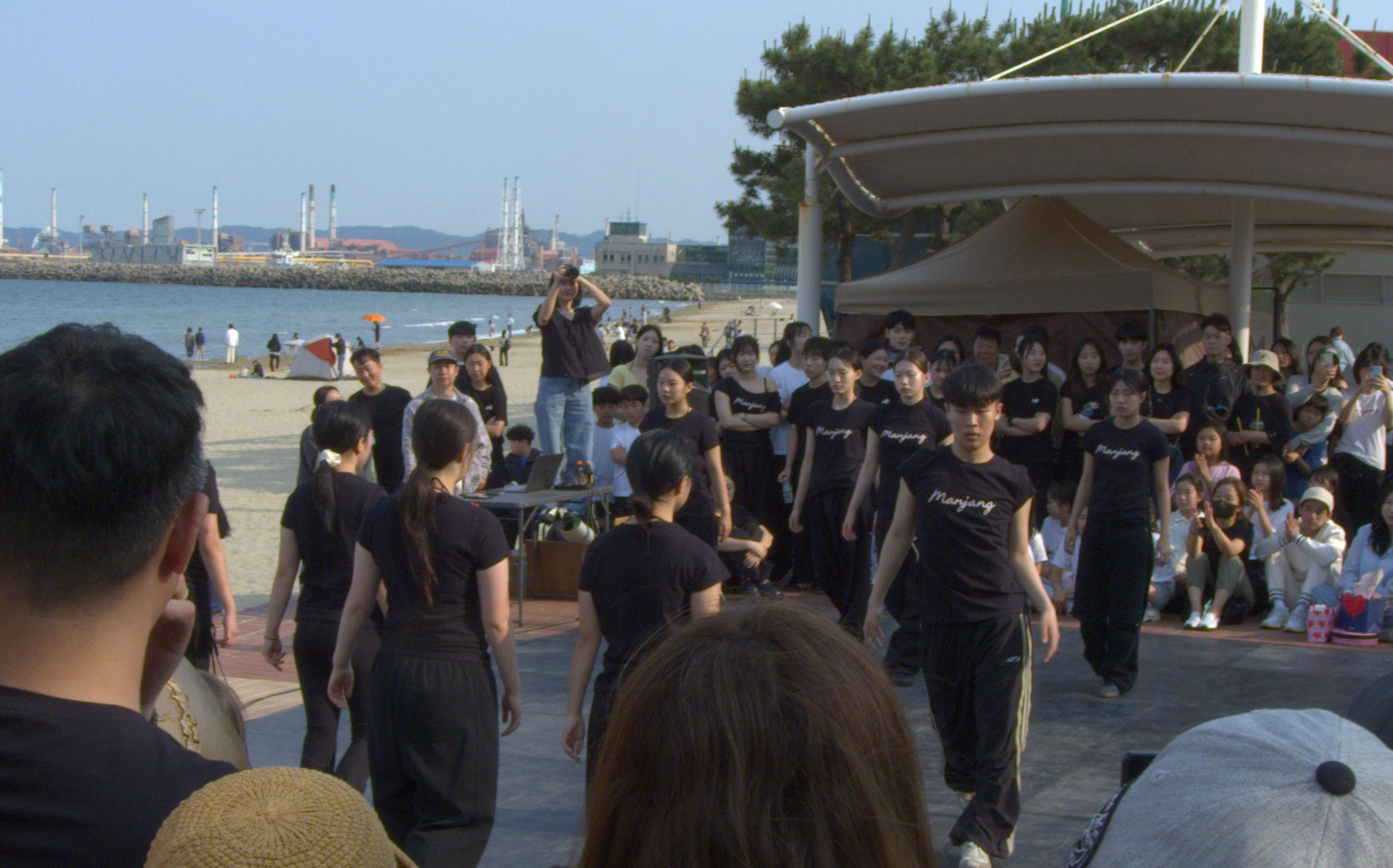}
        % \caption{Caption 2}
    \end{subfigure}
    \begin{subfigure}[b]{0.45\columnwidth}
        \includegraphics[width=\textwidth]{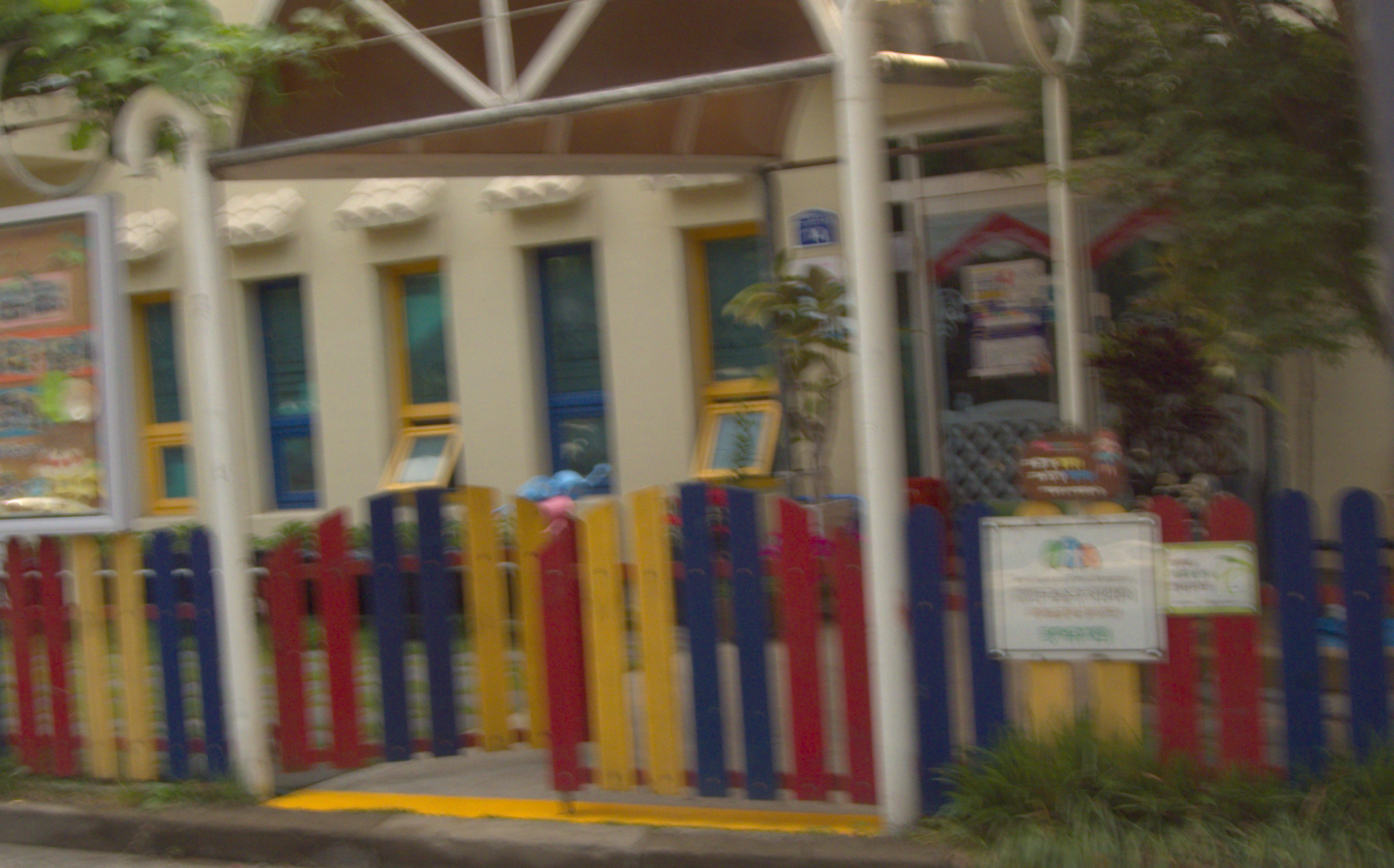}
        % \caption{Caption 3}
    \end{subfigure}
    \begin{subfigure}[b]{0.45\columnwidth}
        \includegraphics[width=\textwidth]{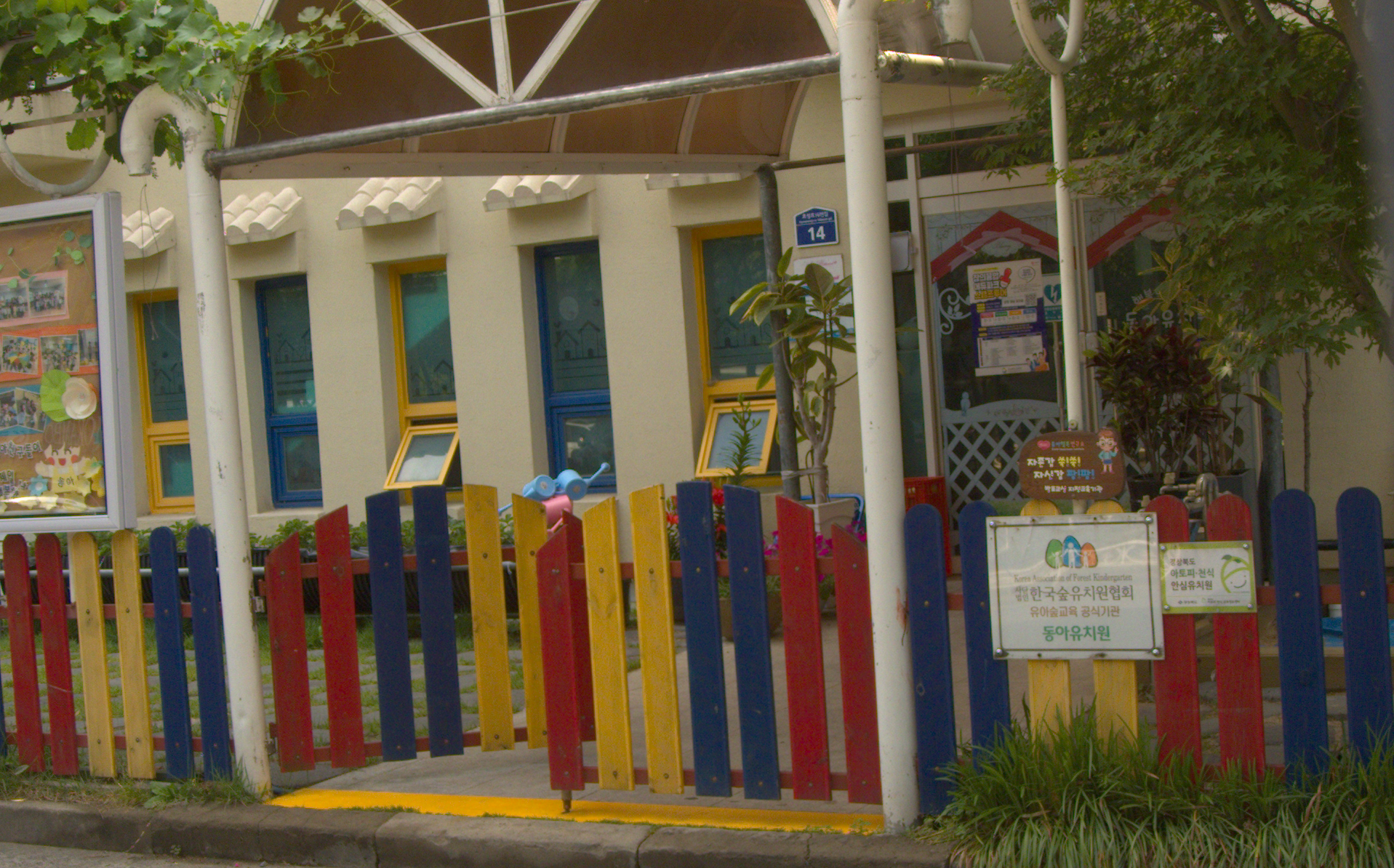}
        % \caption{Caption 4}
    \end{subfigure}
    \caption{Sample of the testset used for the challenge.}
    \label{fig:rsblur_sample}
\end{figure}

For the final testing phase, an extension of the RSBlur dataset was created. 
Using the same system of RSBlur, we collected a new test dataset from scenes entirely disjoint from the original RSBlur dataset.
We also applied the geometrical and photometrical alignment to the test dataset, following the same pipeline of RSBlur.
The new test dataset provides the 420 paired images captured from 84 scenes. A qualitative sample of the test images can be seen in \Cref{fig:rsblur_sample}.

\subsection{Challenge Rules}
\label{subsec:rules}

\begin{table*}[t]
    \centering
    \caption{Final results of the competition. The best and second best results are in \textbf{bold} and \underline{underlined}, respectively.
    }
    \setlength\tabcolsep{2pt}
    \resizebox{0.85\linewidth}{!}{
        \begin{tabular}{r c c c c c c c}
            \toprule
                 Method  & Params$\downarrow$ (M)   & MACs$\downarrow$ (G) & Runtime (ms)     & PSNR$\uparrow$ & SSIM$\uparrow$ & LPIPS$\downarrow$ & Rank  \\\midrule
                 
           NAFRepLocal~(\ref{sec:NAFRepLocal}) & 4.76             & 198.25 & 128.7                  & \textbf{31.130}     & \textbf{0.843} & \underline{0.268} & 1 \\
           
          RestormerL~(\ref{sec:RestormerL}) & \textbf{1.41}             & 199.39 & 149.4            & \underline{31.10}     & \underline{0.840} & 0.280 & 2 \\   
          
          IPIU~(\ref{sec:IPIU}) & 4.35             & \underline{146.33} & 252.8                  & 30.492     & 0.832 &  0.303 & 3   \\ 
          
          SA-NAFNet~(\ref{sec:SA-NAFNet}) & 4.51             & 172.2 & \underline{70.2}                  & 30.189     & 0.819 & \textbf{0.262} & 4 \\ \midrule
          Baseline   & 4.35              & \underline{146.33} & \textbf{63.1}                 & 30.173     & 0.826 & 0.306 & NA \\ 
          Input   &              &  &                 & 26.562     & 0.7091 & 0.326 & \\ \bottomrule
        \end{tabular}
    }
    \label{tab:final-results}
\end{table*} 

\paragraph{Metric} The restored images are compared with the ground truth images. We use a combination of PSNR (Peak-Signal to Noise Ratio), SSIM (Structural Similarity Index), and LPIPS (Learned Perceptual Image Patch Similarity) as the final metric to decide the performance of the solutions. The combination follows the relation:

\begin{equation}
    Score = \lambda_1 \cdot PSNR + \lambda_2 \cdot SSIM + \lambda_3 \cdot LPIPS 
\end{equation}

where $\lambda_1$, $\lambda_2$ and $\lambda_3$ are respectively x, y and z. Due to the geometrical misalignment of the blur-sharp pairs, the restored images could also suffer from misalignment with respect to the sharp images. To reduce the impact of this misalignment, a warp transformation of the blurred image to the sharp one is performed. This raises the metrics results.

\paragraph{Restrictions and Model Parameters and FLOPs} The computational constraints are established on the basis of a study performed on the NAFNet~\cite{chen2022simple} architecture. All the proposed methods need to fulfill them to be valid solutions. We trained different configuration of NAFNet in RSBlur dataset and check which achieved the best efficiency-performance relation. For the sake of comparison, all the configurations were trained following the same pipeline. We random cropped the images into 384x384 crops, applied vertical/horizontal flips as augmentations, and select a batch size of 32. AdamW~\cite{loshchilov2017decoupled} was the optimizer used with weight decay, $\beta_1$ and $\beta_2$ being, respectively, $1e^{-3}$, 0.9 and 0.9. The learning was initialized with a value of $1e^{-3}$ and updated through the cosine annealing strategy~\cite{loshchilov2016sgdr} to a final value of $1^{-6}$. We trained the network for 100 epochs using $\mathcal{L}_1$ loss, using NVIDIA H100 GPUs.

In \Cref{tab:baseline-study}, we present the results of this study. The name of the method is related to the configuration used: C-16 implies an initial embedding of total number of channels 16 and L-28 is the number of blocks in the last encoder step. All other network configuration parameters used are the same as the original NAFNet configuration for GoPro~\cite{chen2022simple}. In addition to the performance results, we show some efficiency parameters calculated for each configuration. The runtimes were calculated by averaging the forward pass of 1000 images of size 1920x1200px, using an RTX 3090 GPU. Multiply-ACCumulate (MACs) operations are calculated also using images of size 1920x1200px. We find the best performance-efficiency relation in the NAFNet-C16-L28 configuration, so we draw the computational restraints based on it: (i) 5M parameters, (ii) 200 GMACs operations.

\paragraph{Others}

The computational restraints must be met by all the proposed methods. In addition, we have limited the use of the RSBlur dataset for training to only the train split. Other external deblurring datasets can be used for training, but need to be noticed in the method report. There are no further constraints in terms of memory occupied or inference time. 

\begin{table}[t]
    \centering
    \caption{Efficiency-Performance study of different NAFNet configurations. %We can clearly see that in this case the single addition was the best possible result.
    }
    \setlength\tabcolsep{2pt}
    \resizebox{1\linewidth}{!}{
        \begin{tabular}{l c c c c c}
            \toprule
                             & Params$\downarrow$ (M)   & MACs$\downarrow$ (G) & Runtime (ms)     & PSNR$\uparrow$ & SSIM$\uparrow$  \\\midrule
          NAFNet-C16-L14 & 2.68             & 94.98 & 52.50                  & 32.23     & 0.836   \\ \midrule
          \textbf{NAFNet-C16-L28} & 4.35             & 146.33 & 61.44            & 32.42     & 0.840 \\ \midrule              
          NAFNet-C24-L14 & 5.98             & 207.93 & 90.50                  & 32.45     & 0.841     \\ \midrule
          NAFNet-C32-L14 & 10.57             & 364.53 & 123.54                  & 32.68     & 0.845  \\ \midrule
          NAFNet-C32-L28   & 17.11              & 566.33 & 151.25                 & 32.83     & 0.848   \\ \bottomrule

        \end{tabular}
    }
    \label{tab:baseline-study}
\end{table} 

\subsection{Challenge Phases}
\label{subsec:phases}

\paragraph{Training and validation phase} Participants are provided with a subset of 100 blurred images randomly selected from the original testing subset of RSBlur. In addition, the RSBlur train split download links are also provided. The training phase lasted for 8 weeks, during which participants could submit their validation results to the CodaBench platform to get PSNR and SSIM feedback. The resulting metrics are displayed on the leaderboard, visible for all participants. The corresponding GT images for the validation set remain hidden from participants, to avoid overfitting of the proposed methods in this phase.

\paragraph{Testing phase} The test phase lasts for three days, during which participants can submit their final results. We provided the participants with 420 blurred images of 84 different scenes, so they could submit the results of their methods to the test leaderbord in CodaBench. Besides, participants are required to send an email to the organizers containing their code and a factsheet, so the organizers can verify and execute the provided code to verify the final results.

\vspace{-2mm}
\paragraph{Related Challenges}
%% cross-referencing AIM 2025 associated challenges
This challenge is one of the AIM 2025~\footnote{\url{https://www.cvlai.net/aim/2025/}} workshop associated challenges on: high FPS non-uniform motion deblurring~\cite{aim2025highfps}, rip current segmentation~\cite{aim2025ripseg}, inverse tone mapping~\cite{aim2025tone}, robust offline video super-resolution~\cite{aim2025videoSR}, low-light raw video denoising~\cite{aim2025videodenoising}, screen-content video quality assessment~\cite{aim2025scvqa}, real-world raw denoising~\cite{aim2025rawdenoising}, 
perceptual image super-resolution~\cite{aim2025perceptual}, 
efficient real-world deblurring~\cite{aim2025efficientdeblurring}, 4K super-resolution on mobile NPUs~\cite{aim20254ksr}, efficient denoising on smartphone GPUs~\cite{aim2025efficientdenoising}, efficient learned ISP on mobile GPUs~\cite{aim2025efficientISP}, and stable diffusion for on-device inference~\cite{aim2025sd}. Descriptions of the datasets, methods, and results can be found in the corresponding challenge reports.
\section{Challenge Results}
\label{subsec:results}

The final results of the competition are listed in \Cref{tab:final-results}. The proposed solutions follow our efficiency criteria: under 5 Million parameters, under 200 GMACs, and run-times bellow 1 second for full-resolution images. The winner of the challenge, MiVideoDeblur (Team Xioami) and 3rd place Team IPIU, propose relevant improvements over NAFNet~\cite{chen2022simple} to improve the efficiency of the method.

\section{Challenge Methods}
\label{subsec:methods}

\begin{disclaimerbox}
In the following Sections, we describe the top challenge solutions -- each was checked manually by the organizers to ensure fairness.

Note that the method descriptions were provided by each team as their contribution to this report.
\end{disclaimerbox}

\newpage
\begin{figure*}[t]
    \centering
    \includegraphics[width=0.8\textwidth]{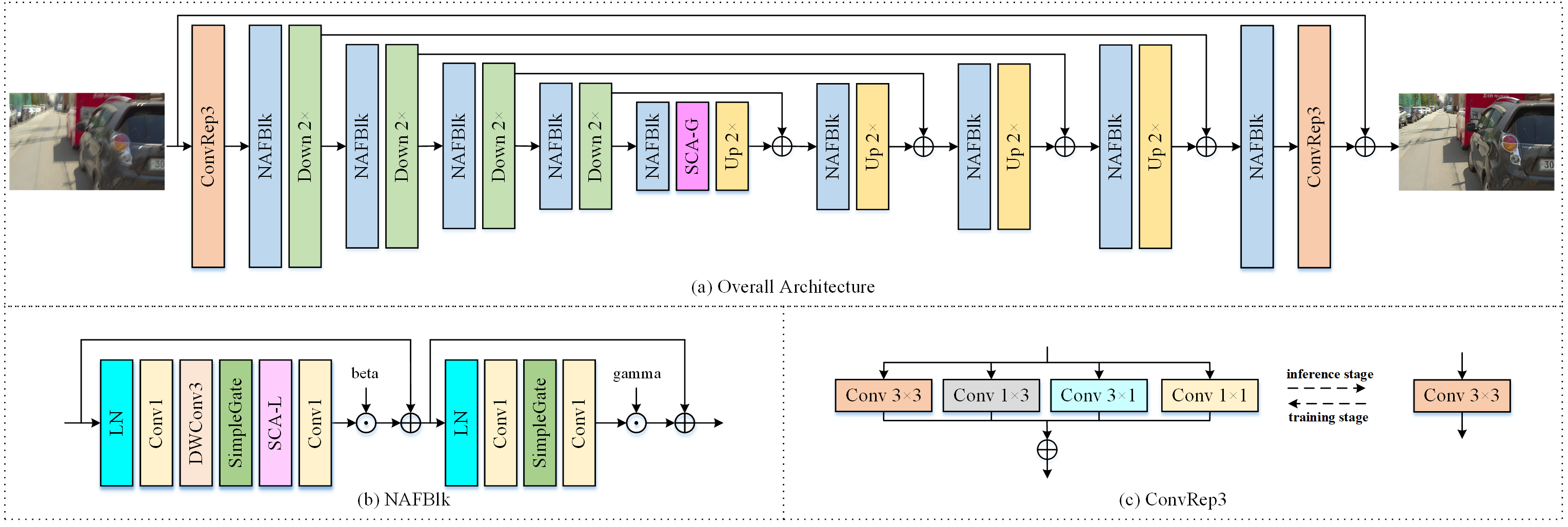}
    \caption{The architecture of NAFRepLocal proposed by MiVideoDeblur.}
    \label{fig:nafreplocal}
\end{figure*}

\subsection{NAFRepLocal}
\label{sec:NAFRepLocal}

%%%%%%%%%%%%%%%%%%%%%%%%%%%%%%%%%%%%
\begin{center}

\vspace{2mm}
\noindent\emph{\textbf{MiVideoDeblur}}
\vspace{2mm}

\noindent\emph{Cheng Li, Jinao Song, Yan Cheng, Long Bao, Heng Sun}

\vspace{2mm}

\noindent\emph{Xiaomi Inc., China}

\vspace{2mm}

\noindent{\emph{Contact: \url{licheng8@xiaomi.com}}}

\end{center}

%%%%%%%%%%%%%%%%%%%%%%%%%%%%%%%%%%%%%%%%%%%%%%%%%%%%%%%%%%%%%%%%%%

\paragraph{Method Description}
This team proposes NAFRepLocal based on NAFNet~\cite{chen2022simple} improvement for single image deblurring, as shown in the Fig~\ref{fig:nafreplocal}. The model as a whole can be divided into 2 layers of convolution, encoding structure, middle structure and decoding structure. The encoding structure contains 4 NAFBlocks and 4 downsampling layers. The number of channels of these 4 NAFBlocks is 32, 64, 128, and 256 respectively. The downsampling layer is implemented by a 2\texttimes2 convolution to achieve 2× downsampling. The number of channels of the NAFBlock in the middle of this model is 512. The decoding structure contains 4 NAFBlocks and 4 upsampling layers. The number of channels of these 4 NAFBlocks are 256, 128, 64 and 32 respectively. The upsampling layer achieves 2× upsampling by a 1\texttimes1 convolution and PixelShuffle. During testing, NAFNet converts the global average pooling in the SCA module (Simplified Channel Attention) into a local operation~\cite{chu2021tlsc}. Here, the former is named SCA-G and the latter is named SCA-L. In contrast, NAFRepLocal directly uses SCA-L module during training. To enhance the model’s ability to extract global features, the team added a SCA-G module after the middle NAFBlock. 

The model only uses the data set provided by the competition and uses reparameterization~\cite{Ding_2019_ICCV} and EMA (Exponential Moving Average) during training. The total number of parameters in the model is about 4.76 M, and the computational effort is about 198.25 GMACs for a 1200 x 1920 input. The solution is illustrated in Figure~\ref{fig:nafreplocal}.

\paragraph{Implementation Details}
The code of this team is based on BasicSR\cite{basicsr}. Before training, each image in the dataset is cropped into patches of 512\texttimes512 and 1024\texttimes1024 sizes, respectively. The model is trained in 4 stages. Reparameterization and SCA-G are only introduced into the model in the fourth stage. At each stage, EMA is used to smooth the weight updates, and the learning rate warm-up is used. The optimizer used during training is AdamW, the learning rate decay strategy is cosine annealing, and the learning rate decays to $1e^{-7}$. The loss function is the PSNR loss. The input is randomly flipped horizontally or vertically during training, and the model is gradually optimized in each stage. A detailed description of the staged training is as follows:

Firstly, the first layer of the model's 3\texttimes3 convolution is replaced by a 5\texttimes5 convolution. The initial learning rate is $2e^{-4}$, the training input is 512\texttimes512, the batch size on each GPU is 8, and 400K steps are trained on 8 A100 GPUs, which takes about 37 hours in total.

Secondly, based on the first stage of training, the training input size is changed to 1024\texttimes1024, and the model is trained for 400K steps on 8 A100 GPUs, which takes about 137 hours in total.

Thirdly, based on the second stage of training, the first convolution layer of the model is replaced with a 3\texttimes3 convolution. The training input size is 1024\texttimes1024, and the model is trained for 400K steps on 8 A100 GPUs, which takes about 132 hours in total.

Lastly, based on the third stage of training, SCA-G is introduced after NAFBlock in the middle of the model, and the first layer and the last layer 3 by 3 are trained using the reparameterization technique. The initial learning rate is $1e^{-4}$, the training input is 1024\texttimes1024, the batch size on each GPU is 4, and 50K steps are trained on 8 H20 GPUs, which takes about 8 hours in total.

This team randomly selected 50 pairs of data from the validation set in the RSBlur dataset as the local validation set. Although the number of training steps in the first three stages was set to 400K, this team actually terminated the training early based on the performance on the local validation set. After four stages of training and optimization of the model structure, the final model training is completed. The model with the highest PSNR on the validation set is finally selected for testing.

\subsection{RestormerL}
\label{sec:RestormerL}

%%%%%%%%%%%%%%%%%%%%%%%%%%%%%%%%%%%%
\begin{center}

\vspace{2mm}
\noindent\emph{\textbf{MAILab}}
\vspace{2mm}

\noindent\emph{Aditya Arora. Akshita Gupta, Anna Rohrbach, Marcus Rohrbach}

\vspace{2mm}

\noindent\emph{TU Durmstadt and hessian.AI, Germany}

\vspace{2mm}

\noindent{\emph{Contact: \url{aditya.arora@tu-darmstadt.de}}}

\end{center}

%%%%%%%%%%%%%%%%%%%%%%%%%%%%%%%%%%%%%%%%%%%%%%%%%%%%%%%%%%%%%%%%%%

\begin{figure*}[t]
\begin{center}
\scalebox{0.85}{
    \begin{tabular}[t]{c} \hspace{-2mm}
    \includegraphics[width=\textwidth]{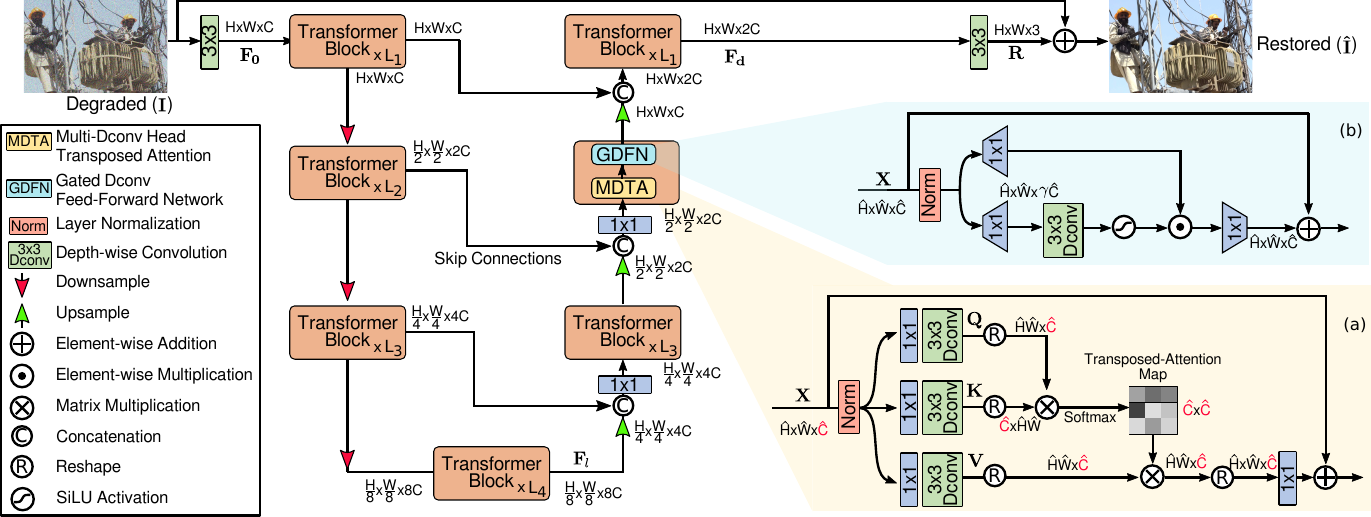}
    \end{tabular}
    }
\end{center}
\caption{\textbf{Overall architecture of RestormerL.} The model employs a symmetric encoder-decoder design with skip connections and utilizes Multi-Dconv Head Transposed Attention (MDTA) and Gated-Dconv Feed-Forward Network (GDFN) blocks. Downsampling and upsampling are achieved via pixel-unshuffle and pixel-shuffle operations, respectively.}

\label{fig:framework}
\end{figure*}

\paragraph{Method Description}
In Figure~\ref{fig:framework} the team presents the overall pipeline of their RestormerL architecture.
Given a noisy image $\mathbf{I}$~$\in$~$\mathbb{R}^{H\times W \times 3}$, RestormerL first applies a convolution to obtain low-level feature embeddings $\mathbf{F_0}$~$\in$~$\mathbb{R}^{H\times W \times C}$; where $H\times W$ denotes the spatial dimension and $C$ is the number of channels. Next, these shallow features $\mathbf{F_0}$ pass through 4-level symmetric encoder-decoder and are transformed into deep features $\mathbf{F_d}$~$\in$~$\mathbb{R}^{H\times W \times C}$.
Starting from the high-resolution input,
the encoder hierarchically reduces spatial size, while expanding channel capacity.
For feature downsampling and upsampling, the team applies pixel-unshuffle and pixel-shuffle operations, respectively.
To assist the recovery process, the encoder features are concatenated with the decoder features via skip connections.
Finally, a convolution layer is applied to the refined features to generate a residual image $\mathbf{R}$~$\in$~$\mathbb{R}^{H\times W \times 3}$. The restored image is then computed by adding the residual to the degraded input, as defined in Equation~\eqref{eq:residual}.

\begin{equation}
    \mathbf{\hat{I}} = \mathbf{I} + \mathbf{R}
\label{eq:residual}
\end{equation}
The model is based on Restormer~\cite{Zamir2021Restormer} architecture, but the team has made some modifications to the original architecture:
 \begin{itemize}
    \item \textbf{Block depth:} Reduced the number of Transformer blocks from \([4, 6, 6, 8]\) to \([2, 2, 2, 4]\).
    \item \textbf{Channel dimensions:} Reduced from \([48, 96, 192, 384]\) to \([16, 32, 64, 128]\).
    \item \textbf{Removed refinement block:} The final refinement stage used in Restormer was omitted.
    \item \textbf{GDFN simplification:} Removed one \(3 \times 3\) depthwise convolution from the gating branch.
    \item \textbf{Activation function:} Replaced GELU with SiLU activation in the GDFN gating mechanism.
    \item \textbf{No pretraining:} The model was trained from scratch using only the RSBlur~\cite{rim2022realistic} dataset, extracting obverlapping patches as a preprocessing technique similar to Restormer.
\end{itemize}

\paragraph{Implementation Details}
The team performs progressive learning where the network is trained on smaller image patches in the early epochs and on gradually larger patches in the later training epochs. The model trained on mixed-size patches via progressive learning shows enhanced performance at test time where images can be of different resolutions (a common case in image restoration).
The model is trained on the RSBlur dataset~\cite{rim2022realistic}. For data augmentation, we use horizontal and vertical flips.

In all experiments, the following training parameters are used. 
From level-1 to level-4, the number of Transformer blocks are [2, 2, 2, 4], attention heads in MDTA are [1, 2, 4, 8], and number of channels are [16, 32, 64, 128]. The refinement stage contains 4 blocks.
The channel expansion factor in GDFN is $\gamma{=}2.2$. 

For progressive learning, they start training with patch size $256$$\times$$256$ and batch size $96$. 
The patch size and batch size pairs are updated to [($384^2$,$64$), ($512^2$,$32$)] at iterations [$100$K, $200$K]. The team uses 8 H100s to train the model during 18h.
The optimizer used is AdamW with $\beta_1{=}0.9$, $\beta_2{=}0.999$ and weight decay${=}1e^{-4}$ hyperparameter values. They use L$_1$ loss for $300$K iterations with the initial learning rate $3e^{-4}$  gradually reduced to $1e^{-6}$ with the cosine annealing.
The whole model is implemented using PyTorch.

\subsection{Enhancing Generalization Capability in Image Deblurring via Data
Augmentation}
\label{sec:IPIU}

%%%%%%%%%%%%%%%%%%%%%%%%%%%%%%%%%%%%
\begin{center}

\vspace{2mm}
\noindent\emph{\textbf{IPIU}}
\vspace{2mm}

\noindent\emph{Meilin Gao, Lianping Lu, Heng Yang}

\vspace{2mm}

\noindent\emph{Intelligent Perception and Image Understanding Lab, Xidian University}

\vspace{2mm}

\noindent{\emph{Contact:} 
\url{24181214076@stu.xidian.edu.cn}, 
\url{24171213904@stu.xidian.edu.cn}, 
\url{24171213995@stu.xidian.edu.cn}}
\end{center}

%%%%%%%%%%%%%%%%%%%%%%%%%%%%%%%%%%%%%%%%%%%%%%%%%%%%%%%%%%%%%%%%%%

\paragraph{Method Description}

\begin{figure}[h]
    \centering
    \begin{subfigure}[b]{0.45\textwidth}
        \includegraphics[width=\textwidth]{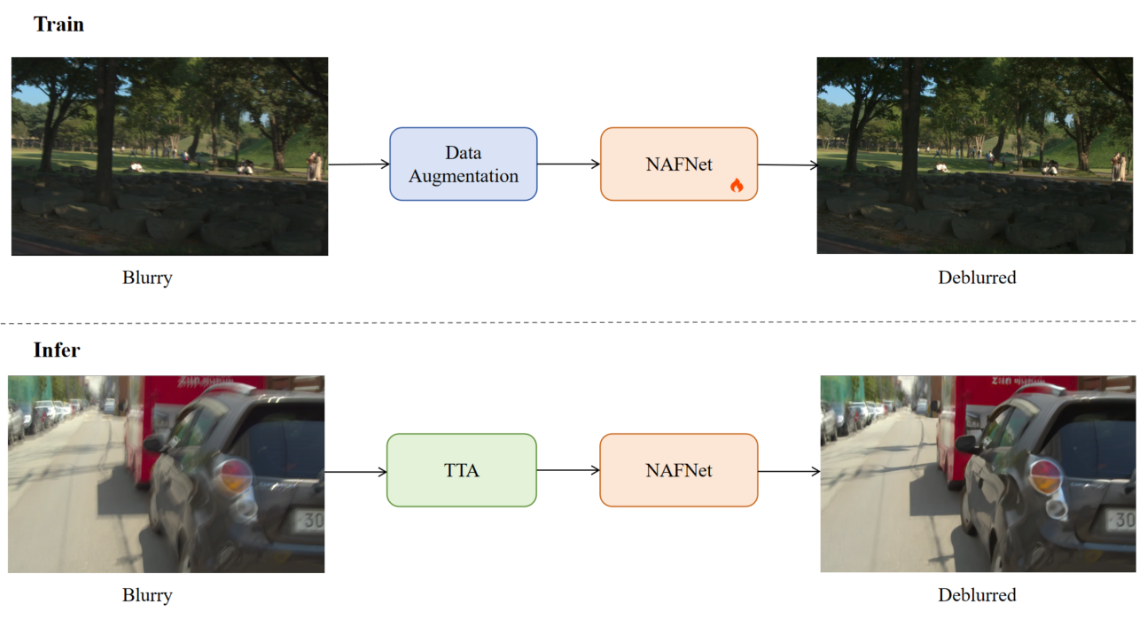} 
        \label{fig:subfig}
    \end{subfigure}
    \caption{The Overview the proposed method architecture by Team IPIU. }
    \label{fig:my_diagram1}
\end{figure}

In this work, the team adopts a novel approach for image deblurring based on an enhanced NAFNet-inspired architecture. The method focuses on efficiently restoring sharp images from blurry ones while preserving fine details and maintaining accurate color information. Figure~\ref{fig:my_diagram1} shows the team proposed method pipeline.

\begin{figure}[t]
    \centering
    \begin{subfigure}[b]{0.1\textwidth}
        \includegraphics[width=\textwidth]{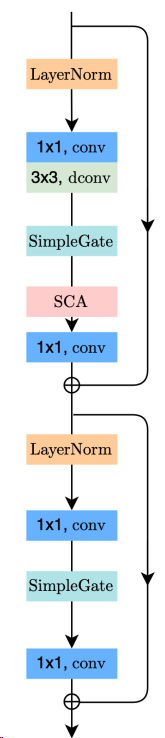} % 
        \caption{NAFNet}
        \label{fig:sub_a}
    \end{subfigure}
    \begin{subfigure}[b]{0.22\textwidth}
        \includegraphics[width=\textwidth]{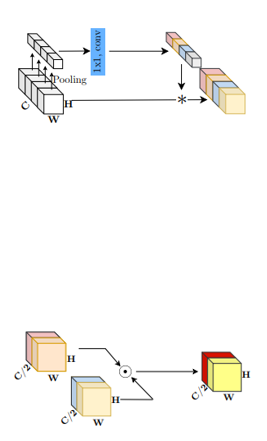}
        \caption{SCA and SG}
        \label{fig:sub_b}
    \end{subfigure}
    \caption{ (a) The architecture of NAFNet, (b) the architecture of SCA and SG.}
    \label{fig:my_diagram}
\end{figure}

Their network architecture is inspired by the NAFNet structure but incorporates modern components for improved performance. As shown in Figure~\ref{fig:sub_a}, NAFNet adopts a single-stage U-shaped architecture and is built on PlainNet, which contains convolution, ReLU~\cite{agarap2018deep}, and shortcut connections. It forms a baseline model by adding Layer Normalization to stabilize training, replacing ReLU with GELU~\cite{lee2023gelu}, and introducing the Channel Attention mechanism. Furthermore, it replaces GELU with SimpleGate, replaces the Channel Attention mechanism with the Simplified Channel Attention, removes all non-linear activation functions, and retains only basic components such as convolution, Layer Normalization, and shortcut.


\paragraph{Implementation Details}
In the training process, the team adopts the NAFNet as the network architecture and use the RSBlur dataset. To enhance training effectiveness, they employ random cropping and scaling augmentation on images, along with horizontal/vertical flipping, as data augmentation strategies. 

The L1 pixel loss is employed as the main loss function, which directly measures the pixel-level differences between the predicted images and the real sharp images, thereby guiding the model to learn basic pixel mapping relationships. Additionally, the authors continuously increase the number of training epochs to achieve optimal performance. 

During inference, Test-Time Augmentation (TTA) is incorporated, including horizontal flipping, vertical flipping, and slight scaling operations (scaling the input tensor down by 10.

The used framework is PyTorch. They use the Adam optimizer with an initial learning rate of \(10^{-3}\), and updated it with the learning rate scheduling strategy. They trained the model on a NVIDIA 3090 GPU for 69 hours.

\subsection{Nonlinear Activation Free Network Improved by Spatial Attention(SA-NAFNet)}
\label{sec:SA-NAFNet}
%%%%%%%%%%%%%%%%%%%%%%%%%%%%%%%%%%%%
\begin{center}

\vspace{2mm}
\noindent\emph{\textbf{XD 2025PBL}}
\vspace{2mm}

\noindent\emph{Bolian Peng,Jianing Liu,Yingxin Wang,}

\vspace{2mm}

%\noindent\emph{Affiliation 1\\Affiliation 2}

\vspace{2mm}

\noindent{\emph{Contact: \url{janeyo0910@gmail.com}}}

\end{center}

%%%%%%%%%%%%%%%%%%%%%%%%%%%%%%%%%%%%%%%%%%%%%%%%%%%%%%%%%%%%%%%%%%
\paragraph{Method Description}

\begin{figure}[h]
       \centering
       \includegraphics[width=1\linewidth]{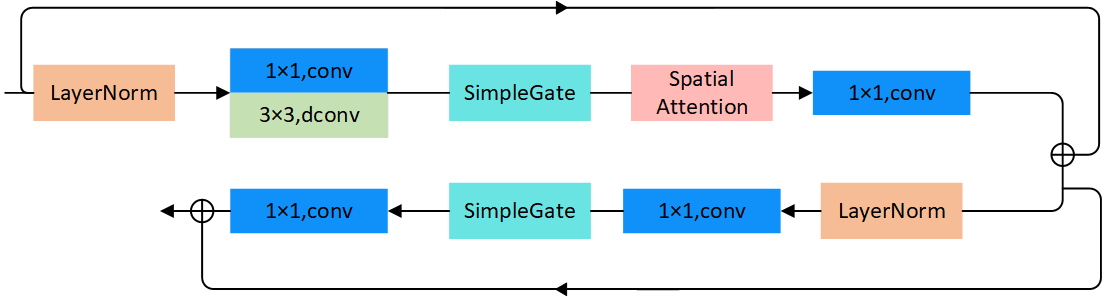}
       \caption{SA-NAFNet Architecture.}
       \label{fig:architecture}
\end{figure}

SA-NAFNet is designed for efficient real-world deblurring, balancing performance and computational cost. It is based on nonlinear activation free network (NAFNet) architecture, integrates spatial attention for better spatial modeling, and uses multi-step training to de-emphasize different focal points in real-world blurring image. Figure~\ref{fig:architecture} shows the architecture proposed by the authors.

For training, the team adopted an end-to-end training approach. They used Adam optimizer and trained for 20 hours on appropriate GPU hardware. The loss function combined pixel loss, perceptual loss and edge loss. During inference, the blurred image is entered, and the model directly outputs the deblurred result, adaptively handling different input sizes.

\paragraph{Implementation details}
The team used PyTorch for the model's implementation. They used the Adam optimizer with an initial learning rate ${=}10^{-3}$, $\beta_1=\beta_2=0.9$, weight decay $=10^{-3}$ and also applied the  CosineAnnealingLR scheduler with $\eta_{min}=10^{-6}$ and $\gamma=0.5$. For efficiency, they also reduced the number of channels.

Their solution used multi-step training across five training steps: 
\begin{itemize}
\item Steps 1-2: Focus on pixel-level details to stabilize spatial attention.
\item Steps 3-4: Introduce perceptual loss and edge loss to enhance structural coherence.
\item Step 5: Balance all losses ($L_1$, perceptual loss, edge loss) for final refinement.
\end{itemize}
The team used 4 NVIDIA 2080Ti and 2 NVIDIA 3090 GPUs and trained for 20 hours.

\newpage
\section{Discussion and Conclusion}
\label{sec:conclusion}

We propose a challenge for real-world image deblurring. Unlike previous academic benchmarks, the proposed dataset and challenge considers two fundamental aspects: real data complexity and efficiency. The proposed solutions can reduce the blur in real high-resolution image (FHD) under 200ms, establishing new baselines for real image deblurring applications.

In the future we plan to host more challenges to promote realistic research and applications in the field of inverse problems and image deblurring, for instance, enforcing more efficiency constraints, or using metadata.

\section*{Acknowledgments}
This work was partially supported by the Alexander von Humboldt Foundation. We thank the AIM 2025 sponsors: AI Witchlabs and University of W\"urzburg (Computer Vision Lab).

Cidaut AI thank Supercomputing of Castile and Leon (SCAYLE. Leon, Spain) for assistance with
the model training and GPU resources.

The organizers thank all the original creators of the RSBlur dataset for their effort and contribution.

{
    \small
    \bibliographystyle{ieeenat_fullname}
    \bibliography{main}

\begin{thebibliography}{44}
\providecommand{\natexlab}[1]{#1}
\providecommand{\url}[1]{\texttt{#1}}
\expandafter\ifx\csname urlstyle\endcsname\relax
  \providecommand{\doi}[1]{doi: #1}\else
  \providecommand{\doi}{doi: \begingroup \urlstyle{rm}\Url}\fi

\bibitem[Agarap(2018)]{agarap2018deep}
Abien~Fred Agarap.
\newblock Deep learning using rectified linear units (relu).
\newblock \emph{arXiv preprint arXiv:1803.08375}, 2018.

\bibitem[Chen et~al.(2022)Chen, Chu, Zhang, and Sun]{chen2022simple}
Liangyu Chen, Xiaojie Chu, Xiangyu Zhang, and Jian Sun.
\newblock Simple baselines for image restoration.
\newblock In \emph{European conference on computer vision}, pages 17--33. Springer, 2022.

\bibitem[Cho et~al.(2021)Cho, Ji, Hong, Jung, and Ko]{Cho2021}
Sung-Jin Cho, Seo-Won Ji, Jun-Pyo Hong, Seung-Won Jung, and Sung-Jea Ko.
\newblock Rethinking coarse-to-fine approach in single image deblurring.
\newblock In \emph{Proceedings of the IEEE/CVF International Conference on Computer Vision (ICCV)}, 2021.

\bibitem[Chu et~al.(2021)Chu, Chen, , Chen, and Lu]{chu2021tlsc}
Xiaojie Chu, Liangyu Chen, , Chengpeng Chen, and Xin Lu.
\newblock Revisiting global statistics aggregation for improving image restoration.
\newblock \emph{arXiv preprint arXiv:2112.04491}, 2021.

\bibitem[Ciubotariu et~al.(2025)Ciubotariu, Vasluianu, Zhou, Mehta, Timofte, et~al.]{aim2025highfps}
George Ciubotariu, Florin-Alexandru Vasluianu, Zhuyun Zhou, Nancy Mehta, Radu Timofte, et~al.
\newblock {AIM} 2025 high {FPS} non-uniform motion deblurring challenge report.
\newblock In \emph{Proceedings of the IEEE/CVF International Conference on Computer Vision (ICCV) Workshops}, 2025.

\bibitem[Conde et~al.(2023)Conde, Zamfir, Timofte, Motilla, et~al.]{Conde_2023_CVPR}
Marcos~V. Conde, Eduard Zamfir, Radu Timofte, Daniel Motilla, et~al.
\newblock Efficient deep models for real-time 4k image super-resolution. ntire 2023 benchmark and report.
\newblock In \emph{Proceedings of the IEEE/CVF Conference on Computer Vision and Pattern Recognition (CVPR) Workshops}, pages 1495--1521, 2023.

\bibitem[Ding et~al.(2019)Ding, Guo, Ding, and Han]{Ding_2019_ICCV}
Xiaohan Ding, Yuchen Guo, Guiguang Ding, and Jungong Han.
\newblock Acnet: Strengthening the kernel skeletons for powerful cnn via asymmetric convolution blocks.
\newblock In \emph{The IEEE International Conference on Computer Vision (ICCV)}, 2019.

\bibitem[Dumitriu et~al.(2025)Dumitriu, Miron, Tatui, Ionescu, Timofte, Ralhan, Vasluianu, et~al.]{aim2025ripseg}
Andrei Dumitriu, Florin Miron, Florin Tatui, Radu~Tudor Ionescu, Radu Timofte, Aakash Ralhan, Florin-Alexandru Vasluianu, et~al.
\newblock {AIM} 2025 challenge on rip current segmentation ({RipSeg}).
\newblock In \emph{Proceedings of the IEEE/CVF International Conference on Computer Vision (ICCV) Workshops}, 2025.

\bibitem[Feijoo et~al.(2025{\natexlab{a}})Feijoo, Benito, Garcia, and Conde]{Feijoo_2025_CVPR}
Daniel Feijoo, Juan~C. Benito, Alvaro Garcia, and Marcos~V. Conde.
\newblock Darkir: Robust low-light image restoration.
\newblock In \emph{Proceedings of the Computer Vision and Pattern Recognition Conference (CVPR)}, pages 10879--10889, 2025{\natexlab{a}}.

\bibitem[Feijoo et~al.(2025{\natexlab{b}})Feijoo, Garrido, Conde, Rim, Garcia, Cho, Timofte, et~al.]{aim2025efficientdeblurring}
Daniel Feijoo, Paula Garrido, Marcos Conde, Jaesung Rim, Alvaro Garcia, Sunghyun Cho, Radu Timofte, et~al.
\newblock Efficient real-world deblurring using single images: {AIM} 2025 challenge report.
\newblock In \emph{Proceedings of the IEEE/CVF International Conference on Computer Vision (ICCV) Workshops}, 2025{\natexlab{b}}.

\bibitem[Ignatov et~al.(2025{\natexlab{a}})Ignatov, Perevozchikov, Timofte, et~al.]{aim20254ksr}
Andrey Ignatov, Georgy Perevozchikov, Radu Timofte, et~al.
\newblock {4K} image super-resolution on mobile {NPUs}: {Mobile AI \& AIM 2025} challenge report.
\newblock In \emph{Proceedings of the IEEE/CVF International Conference on Computer Vision (ICCV) Workshops}, 2025{\natexlab{a}}.

\bibitem[Ignatov et~al.(2025{\natexlab{b}})Ignatov, Perevozchikov, Timofte, et~al.]{aim2025efficientISP}
Andrey Ignatov, Georgy Perevozchikov, Radu Timofte, et~al.
\newblock Efficient learned smartphone {ISP} on mobile {GPUs}: {Mobile AI \& AIM 2025} challenge report.
\newblock In \emph{Proceedings of the IEEE/CVF International Conference on Computer Vision (ICCV) Workshops}, 2025{\natexlab{b}}.

\bibitem[Ignatov et~al.(2025{\natexlab{c}})Ignatov, Perevozchikov, Timofte, et~al.]{aim2025efficientdenoising}
Andrey Ignatov, Georgy Perevozchikov, Radu Timofte, et~al.
\newblock Efficient image denoising on smartphone {GPUs}: {Mobile AI \& AIM 2025} challenge report.
\newblock In \emph{Proceedings of the IEEE/CVF International Conference on Computer Vision (ICCV) Workshops}, 2025{\natexlab{c}}.

\bibitem[Ignatov et~al.(2025{\natexlab{d}})Ignatov, Perevozchikov, Timofte, et~al.]{aim2025sd}
Andrey Ignatov, Georgy Perevozchikov, Radu Timofte, et~al.
\newblock Adapting stable diffusion for on-device inference: {Mobile AI \& AIM 2025} challenge report.
\newblock In \emph{Proceedings of the IEEE/CVF International Conference on Computer Vision (ICCV) Workshops}, 2025{\natexlab{d}}.

\bibitem[Jiang et~al.(2024)Jiang, Zhang, Gao, and Deng]{sfhformer}
Xingyu Jiang, Xiuhui Zhang, Ning Gao, and Yue Deng.
\newblock When fast fourier transform meets transformer for image restoration.
\newblock In \emph{European Conference on Computer Vision}, pages 381--402. Springer, 2024.

\bibitem[Karetin et~al.(2025)Karetin, Molodetskikh, Vatolin, Timofte, et~al.]{aim2025videoSR}
Nikolai Karetin, Ivan Molodetskikh, Dmitry Vatolin, Radu Timofte, et~al.
\newblock {AIM} 2025 challenge on robust offline video super-resolution: Dataset, methods and results.
\newblock In \emph{Proceedings of the IEEE/CVF International Conference on Computer Vision (ICCV) Workshops}, 2025.

\bibitem[Kong et~al.(2023)Kong, Dong, Ge, Li, and Pan]{kong2023efficient}
Lingshun Kong, Jiangxin Dong, Jianjun Ge, Mingqiang Li, and Jinshan Pan.
\newblock Efficient frequency domain-based transformers for high-quality image deblurring.
\newblock In \emph{Proceedings of the IEEE/CVF Conference on Computer Vision and Pattern Recognition}, pages 5886--5895, 2023.

\bibitem[Lee(2023)]{lee2023gelu}
Minhyeok Lee.
\newblock Gelu activation function in deep learning: a comprehensive mathematical analysis and performance.
\newblock \emph{arXiv preprint arXiv:2305.12073}, 2023.

\bibitem[Lee et~al.(2025)Lee, Park, Canelo, Park, Kim, Chun, Jin, Li, Guo, Timofte, et~al.]{lee2025ntire}
Sangmin Lee, Eunpil Park, Angel Canelo, Hyunhee Park, Youngjo Kim, Hyungju Chun, Xin Jin, Chongyi Li, Chun-Le Guo, Radu Timofte, et~al.
\newblock Ntire 2025 challenge on efficient burst hdr and restoration: Datasets, methods, and results.
\newblock In \emph{Proceedings of the Computer Vision and Pattern Recognition Conference}, pages 1002--1017, 2025.

\bibitem[Li et~al.(2025)Li, Li, Conde, Besbinar, Hosu, Iso, Timofte, et~al.]{aim2025rawdenoising}
Feiran Li, Jiacheng Li, Marcos Conde, Beril Besbinar, Vlad Hosu, Daisuke Iso, Radu Timofte, et~al.
\newblock Real-world raw denoising using diverse cameras: {AIM} 2025 challenge report.
\newblock In \emph{Proceedings of the IEEE/CVF International Conference on Computer Vision (ICCV) Workshops}, 2025.

\bibitem[Li et~al.(2023)Li, Zhang, Jiang, Luo, Feng, and Xu]{Li2023}
Haoying Li, Ziran Zhang, Tingting Jiang, Peng Luo, Huajun Feng, and Zhihai Xu.
\newblock Real-world deep local motion deblurring.
\newblock In \emph{proceedings of the AAAI conference on artificial intelligence}, pages 1314--1322, 2023.

\bibitem[Longarela et~al.(2025)Longarela, Conde, Álvaro García, Timofte, et~al.]{aim2025perceptual}
Bruno Longarela, Marcos Conde, Álvaro García, Radu Timofte, et~al.
\newblock {AIM} 2025 perceptual image super-resolution challenge.
\newblock In \emph{Proceedings of the IEEE/CVF International Conference on Computer Vision (ICCV) Workshops}, 2025.

\bibitem[Loshchilov and Hutter(2016)]{loshchilov2016sgdr}
Ilya Loshchilov and Frank Hutter.
\newblock Sgdr: Stochastic gradient descent with warm restarts.
\newblock \emph{arXiv preprint arXiv:1608.03983}, 2016.

\bibitem[Loshchilov and Hutter(2017)]{loshchilov2017decoupled}
Ilya Loshchilov and Frank Hutter.
\newblock Decoupled weight decay regularization.
\newblock \emph{arXiv preprint arXiv:1711.05101}, 2017.

\bibitem[Nah et~al.(2017)Nah, Kim, and Lee]{Nah2017}
Seungjun Nah, Tae~Hyun Kim, and Kyoung~Mu Lee.
\newblock Deep multi-scale convolutional neural network for dynamic scene deblurring.
\newblock In \emph{Proceedings of the IEEE Conference on Computer Vision and Pattern Recognition (CVPR)}, 2017.

\bibitem[Nah et~al.(2019{\natexlab{a}})Nah, Baik, Hong, Moon, Son, Timofte, and Mu~Lee]{nah2019ntire_sr}
Seungjun Nah, Sungyong Baik, Seokil Hong, Gyeongsik Moon, Sanghyun Son, Radu Timofte, and Kyoung Mu~Lee.
\newblock Ntire 2019 challenge on video deblurring and super-resolution: Dataset and study.
\newblock In \emph{Proceedings of the IEEE/CVF conference on computer vision and pattern recognition workshops}, pages 0--0, 2019{\natexlab{a}}.

\bibitem[Nah et~al.(2019{\natexlab{b}})Nah, Timofte, Baik, Hong, Moon, Son, and Mu~Lee]{nah2019ntire}
Seungjun Nah, Radu Timofte, Sungyong Baik, Seokil Hong, Gyeongsik Moon, Sanghyun Son, and Kyoung Mu~Lee.
\newblock Ntire 2019 challenge on video deblurring: Methods and results.
\newblock In \emph{Proceedings of the IEEE/CVF Conference on Computer Vision and Pattern Recognition Workshops}, pages 0--0, 2019{\natexlab{b}}.

\bibitem[Nah et~al.(2020)Nah, Son, Timofte, and Lee]{nah2020ntire}
Seungjun Nah, Sanghyun Son, Radu Timofte, and Kyoung~Mu Lee.
\newblock Ntire 2020 challenge on image and video deblurring.
\newblock In \emph{Proceedings of the IEEE/CVF Conference on Computer Vision and Pattern Recognition Workshops}, pages 416--417, 2020.

\bibitem[Nah et~al.(2021)Nah, Son, Lee, Timofte, Lee, Chen, Zhang, Lu, Chu, Chen, et~al.]{nah2021ntire}
Seungjun Nah, Sanghyun Son, Suyoung Lee, Radu Timofte, Kyoung~Mu Lee, Liangyu Chen, Jie Zhang, Xin Lu, Xiaojie Chu, Chengpeng Chen, et~al.
\newblock Ntire 2021 challenge on image deblurring.
\newblock In \emph{Proceedings of the IEEE/CVF Conference on Computer Vision and Pattern Recognition}, pages 149--165, 2021.

\bibitem[Ren et~al.(2024)Ren, Li, Mehta, Timofte, Yu, Wan, Hong, Han, Wu, Zou, et~al.]{ren2024ninth}
Bin Ren, Yawei Li, Nancy Mehta, Radu Timofte, Hongyuan Yu, Cheng Wan, Yuxin Hong, Bingnan Han, Zhuoyuan Wu, Yajun Zou, et~al.
\newblock The ninth ntire 2024 efficient super-resolution challenge report.
\newblock In \emph{Proceedings of the IEEE/CVF Conference on Computer Vision and Pattern Recognition}, pages 6595--6631, 2024.

\bibitem[Ren et~al.(2025)Ren, Guo, Sun, Wu, Timofte, and Li]{ren2025tenth}
Bin Ren, Hang Guo, Lei Sun, Zongwei Wu, Radu Timofte, and Yawei Li.
\newblock The tenth ntire 2025 efficient super-resolution challenge report.
\newblock In \emph{Proceedings of the Computer Vision and Pattern Recognition Conference}, pages 917--966, 2025.

\bibitem[Rim et~al.(2020)Rim, Lee, Won, and Cho]{realblur}
Jaesung Rim, Haeyun Lee, Jucheol Won, and Sunghyun Cho.
\newblock Real-world blur dataset for learning and benchmarking deblurring algorithms.
\newblock In \emph{Computer vision--ECCV 2020: 16th European conference, glasgow, UK, August 23--28, 2020, proceedings, part XXV 16}, pages 184--201. Springer, 2020.

\bibitem[Rim et~al.(2022)Rim, Kim, Kim, Lee, Lee, and Cho]{rim2022realistic}
Jaesung Rim, Geonung Kim, Jungeon Kim, Junyong Lee, Seungyong Lee, and Sunghyun Cho.
\newblock Realistic blur synthesis for learning image deblurring.
\newblock In \emph{European conference on computer vision}, pages 487--503. Springer, 2022.

\bibitem[Safonov et~al.(2025)Safonov, Rakhmanov, Vatolin, Timofte, et~al.]{aim2025scvqa}
Nickolay Safonov, Mikhail Rakhmanov, Dmitriy Vatolin, Radu Timofte, et~al.
\newblock {AIM} 2025 challenge on screen-content video quality assessment: Methods and results.
\newblock In \emph{Proceedings of the IEEE/CVF International Conference on Computer Vision (ICCV) Workshops}, 2025.

\bibitem[Su et~al.(2017)Su, Delbracio, Wang, Sapiro, Heidrich, and Wang]{su2017deep}
Shuochen Su, Mauricio Delbracio, Jue Wang, Guillermo Sapiro, Wolfgang Heidrich, and Oliver Wang.
\newblock Deep video deblurring for hand-held cameras.
\newblock In \emph{Proceedings of the IEEE Conference on Computer Vision and Pattern Recognition}, pages 1279--1288, 2017.

\bibitem[Tsai et~al.(2022{\natexlab{a}})Tsai, Peng, Lin, Tsai, and Lin]{tsai2022stripformer}
Fu-Jen Tsai, Yan-Tsung Peng, Yen-Yu Lin, Chung-Chi Tsai, and Chia-Wen Lin.
\newblock Stripformer: Strip transformer for fast image deblurring.
\newblock In \emph{European Conference on Computer Vision}, pages 146--162. Springer, 2022{\natexlab{a}}.

\bibitem[Tsai et~al.(2022{\natexlab{b}})Tsai, Peng, Tsai, Lin, and Lin]{tsai2022banet}
Fu-Jen Tsai, Yan-Tsung Peng, Chung-Chi Tsai, Yen-Yu Lin, and Chia-Wen Lin.
\newblock Banet: a blur-aware attention network for dynamic scene deblurring.
\newblock \emph{IEEE Transactions on Image Processing}, 31:\penalty0 6789--6799, 2022{\natexlab{b}}.

\bibitem[Wang et~al.(2025)Wang, Banterle, Ren, Timofte, et~al.]{aim2025tone}
Chao Wang, Francesco Banterle, Bin Ren, Radu Timofte, et~al.
\newblock {AIM} 2025 challenge on inverse tone mapping report: Methods and results.
\newblock In \emph{Proceedings of the IEEE/CVF International Conference on Computer Vision (ICCV) Workshops}, 2025.

\bibitem[Wang et~al.(2022)Wang, Xie, Yu, Chan, Loy, and Dong]{basicsr}
Xintao Wang, Liangbin Xie, Ke Yu, Kelvin~C.K. Chan, Chen~Change Loy, and Chao Dong.
\newblock {BasicSR}: Open source image and video restoration toolbox.
\newblock \url{https://github.com/XPixelGroup/BasicSR}, 2022.

\bibitem[Yakovenko et~al.(2025)Yakovenko, Chakvetadze, Khrapov, Zhelezov, Vatolin, Timofte, et~al.]{aim2025videodenoising}
Alexander Yakovenko, George Chakvetadze, Ilya Khrapov, Maksim Zhelezov, Dmitry Vatolin, Radu Timofte, et~al.
\newblock {AIM} 2025 low-light raw video denoising challenge: Dataset, methods and results.
\newblock In \emph{Proceedings of the IEEE/CVF International Conference on Computer Vision (ICCV) Workshops}, 2025.

\bibitem[Zamfir et~al.(2023)Zamfir, Conde, and Timofte]{Zamfir_2023_CVPR}
Eduard Zamfir, Marcos~V. Conde, and Radu Timofte.
\newblock Towards real-time 4k image super-resolution.
\newblock In \emph{Proceedings of the IEEE/CVF Conference on Computer Vision and Pattern Recognition (CVPR) Workshops}, pages 1522--1532, 2023.

\bibitem[Zamir et~al.(2022)Zamir, Arora, Khan, Hayat, Khan, and Yang]{Zamir2021Restormer}
Syed~Waqas Zamir, Aditya Arora, Salman Khan, Munawar Hayat, Fahad~Shahbaz Khan, and Ming-Hsuan Yang.
\newblock Restormer: Efficient transformer for high-resolution image restoration.
\newblock In \emph{CVPR}, 2022.

\bibitem[Zhong et~al.(2020)Zhong, Gao, Zheng, and Zheng]{zhong2020efficient}
Zhihang Zhong, Ye Gao, Yinqiang Zheng, and Bo Zheng.
\newblock Efficient spatio-temporal recurrent neural network for video deblurring.
\newblock In \emph{Computer Vision--ECCV 2020: 16th European Conference, Glasgow, UK, August 23--28, 2020, Proceedings, Part VI 16}, pages 191--207. Springer, 2020.

\bibitem[Zhong et~al.(2023)Zhong, Gao, Zheng, Zheng, and Sato]{zhong2023real}
Zhihang Zhong, Ye Gao, Yinqiang Zheng, Bo Zheng, and Imari Sato.
\newblock Real-world video deblurring: A benchmark dataset and an efficient recurrent neural network.
\newblock \emph{International Journal of Computer Vision}, 131\penalty0 (1):\penalty0 284--301, 2023.

\end{thebibliography}
}

\end{document}